\documentclass[10pt,twocolumn,letterpaper]{article}

\usepackage{iccv}
\usepackage{times}
\usepackage{epsfig}
\usepackage{graphicx}
\usepackage{amsmath}
\usepackage{amssymb}
\usepackage[flushleft]{threeparttable}
\usepackage{multicol}
\usepackage{multirow}
\usepackage{booktabs}
\usepackage{bbm}
\usepackage{subcaption}
\usepackage{stackengine}
\usepackage[export]{adjustbox}
\usepackage{array}
\pdfoutput=1
\newcommand{\comment}[1]{\ignorespaces}

\usepackage[breaklinks=true,bookmarks=false]{hyperref}
\hypersetup{
    colorlinks=true,
    linkcolor=blue,
    filecolor=magenta,      
    urlcolor=cyan,
    }
\iccvfinalcopy 

\def\iccvPaperID{8571} 

\ificcvfinal\fi

\begin{document}

\title{Fine-grained Semantics-aware Representation Enhancement for Self-supervised Monocular Depth Estimation}

\author{Hyunyoung Jung\\
Seoul National University\\
{\tt\small gusdud1500@gmail.com}
\and
Eunhyeok Park \\
POSTECH\\
{\tt\small eh.park@postech.ac.kr}
\and
Sungjoo Yoo\\
Seoul National University\\
{\tt\small sungjoo.yoo@gmail.com}
}

\maketitle
\ificcvfinal\thispagestyle{empty}\fi

 
\begin{abstract}
Self-supervised monocular depth estimation has been widely studied, owing to its practical importance and recent promising improvements. However, most works suffer from limited supervision of photometric consistency, especially in weak texture regions and at object boundaries. To overcome this weakness, we propose novel ideas to improve self-supervised monocular depth estimation by leveraging cross-domain information, especially scene semantics. We focus on incorporating implicit semantic knowledge into geometric representation enhancement and suggest two ideas: a metric learning approach that exploits the semantics-guided local geometry to optimize intermediate depth representations and a novel feature fusion module that judiciously utilizes cross-modality between two heterogeneous feature representations. We comprehensively evaluate our methods on the KITTI dataset and demonstrate that our method outperforms state-of-the-art methods. The source code is available at \url{https://github.com/hyBlue/FSRE-Depth}.
\end{abstract}

\section{Introduction}
Depth measurement is a critical task in various applications, including robotics, augmented reality, and self-driving vehicles. It measures the distance from all or a part of the pixels in the imaging device to target objects using active/passive sensors. Equipping such devices requires high cost and continuous operation, which makes its use limited. Monocular depth estimation estimates the depth of pixels in a given 2D image without additional measurement. It facilitates the understanding of 3D scene geometry from a captured image, which closes the dimension gap between the physical world and an image. 

Because of its importance and cost benefits, there have been lots of studies~\cite{stereo,sfm1,3d,sfm2,10.1145/2601097.2601165} that have improved depth estimation accuracy, temporal consistency and depth ranges. Owing to the success of the convolutional neural network, it has also been adapted to monocular depth estimation and has produced great improvements.

\begin{figure}[t!]
\centering
 \captionsetup[subfigure]{labelformat=empty}
    \footnotesize
    
\subfloat[]{
   \raisebox{0.2in}{\rotatebox[origin=t]{90}{Image}}
   \hspace{-1.975mm}
}
 \subfloat[]{
   \includegraphics[width=0.95\linewidth]{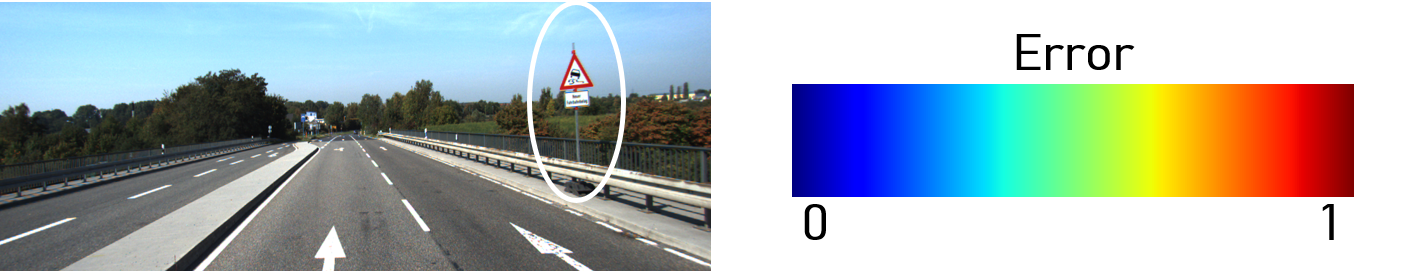}
}
\hfill
\vspace{-6.9mm}

\subfloat[]{
   \raisebox{0.2in}{\rotatebox[origin=t]{90}{\cite{Klingner2020SelfSupervisedMD}}  
   \hspace{-1.85mm}
}}
 \subfloat[]{
   \includegraphics[width=0.95\linewidth]{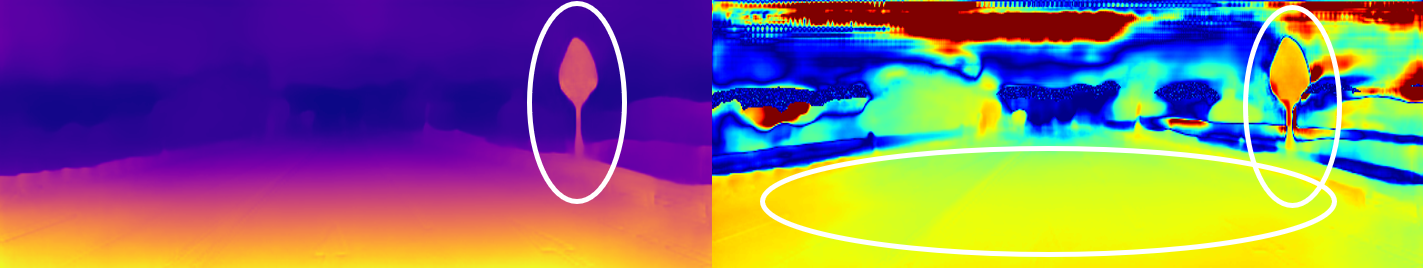}
}
\hfill
\vspace{-6.5mm}

\subfloat[]{
   \raisebox{0.2in}{\rotatebox[origin=t]{90}{\cite{Guizilini20203DPF}}  
      \hspace{-1.85mm}
}}
 \subfloat[]{
   \includegraphics[width=0.95\linewidth]{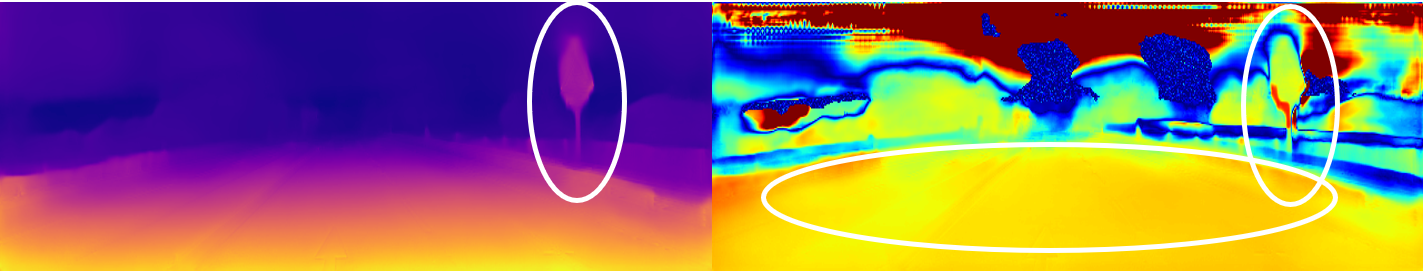}
}
\hfill
\vspace{-6.5mm}

\subfloat[]{
   \raisebox{0.2in}{\rotatebox[origin=t]{90}{Ours}}  
      \hspace{-1.6mm}
}
 \subfloat[]{
   \includegraphics[width=0.95\linewidth]{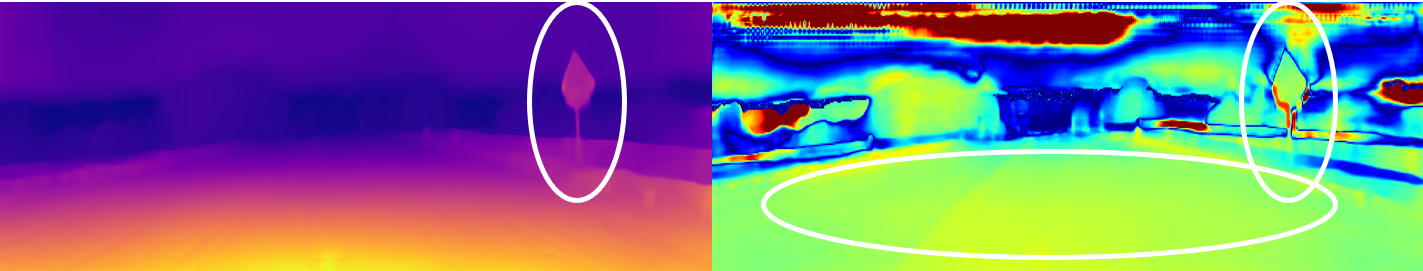}
}
    \vspace{-8mm}
    \caption{Depth predictions and error maps of recent state-of-the-arts~\cite{Guizilini20203DPF, Klingner2020SelfSupervisedMD}. Apart from ours, both lack accuracy in low-texture regions and object boundaries, owing to weak supervision using contemporary self-supervised training.} 
    \vspace{-1.5mm}
    \label{fig:intro}
\end{figure}

Many existing monocular depth estimation methods train their networks with supervised depth labels computed via synthetic data or estimated from depth sensor~\cite{Liu01,Eigen01,Li01,Laina01}. Although such methods have provided significant improvements in depth estimation, they still have multiple concerns related to the high cost of labeling and obtaining the depth labels on pixels, the limited available ground-truth depth data, the restricted depth range of sampled data, and the noticeable noise in the depth values. To avoid these shortcomings, self-supervised training methods have recently been proposed.

Notably, the SfM-Learner~\cite{Zhou01} method utilizes the ensembles of consecutive frames in video sequences for joint training depth and pose networks. It demonstrates comparable performance to extant supervised methods; however, recent works~\cite{dfnet,CC,Godard02,Bian01} based on SfM-Learner mostly rely on photometric loss~\cite{ssim} and smoothness constraints; hence, they suffer from limited supervision of weak texture regions. Furthermore, moving objects and uncertainty in the pose network destabilize training, leading to incorrect depth values, especially on object boundaries (see Fig.~\ref{fig:intro}).

Several recent methods have attempted to overcome this weakness by employing cross-domain knowledge learning, including leveraging scene semantics to improve monocular depth predictions~\cite{Klingner2020SelfSupervisedMD, lee2021learning, wild, casser2018depth}. They remove dynamic objects or explicitly model the object motion from the semantic instances to incorporate them into the scene geometry. In addition, a regularization of the depth smoothness within corresponding semantic objects enforces consistency between depth and semantic predictions~\cite{Chen01, Ramirez01, zhu2020edge}.

In this study, we aim to improve self-supervised monocular depth estimation via the implicit use of semantic segmentation. We do not explicitly identify moving objects or regularize depth values in accordance with the semantic labels. Instead, we focus on representation enhancement, optimizing the depth network in the representation spaces, to produce semantically consistent intermediate depth representations. 

Inspired by the recent use of deep metric learning~\cite{Wang_2017_ICCV, song2017deep, MLarticle}, we suggest a novel semantics-guided triplet loss to refine depth representations according to implicit semantic guidance. Here, our goal is to take advantage of local geometric information from the scene semantics. For example, the adjacent pixels within each object have similar depth values, whereas those that cross semantic boundaries may have large differences. Combined with a simple but effective patch-based sampling strategy, our metric-learning approach exploits the semantics-guided local geometry information to optimize pixel representations near the object boundary, thereby yielding improved depth predictions.

We also design a cross-task attention module for refining depth features more semantically consistent. It computes the similarity between the reference and target features through multiple representation subspaces and effectively utilizes the cross-modal interactions among the heterogeneous representations. As a result, we quantify the semantic awareness of depth features as a form of attention and exploit it to produce  better depth predictions.


Our contributions are summarized as follows. First, we present a novel training method that extracts semantics-guided local geometry with patch-based sampling and utilizes it to refine depth features in a metric-learning formulation. Second, we propose a new cross-task feature fusion architecture that fully utilizes the implicit representations of semantics for learning depth features. Finally, we comprehensively evaluate the performances of these two methods using the KITTI Eigen split and demonstrate that our method outperforms recent state-of-the-art self-supervised monocular depth prediction works in every metric.

\section{Related Work}
\subsection{Depth Estimation with Neural Network}
The recent success of neural networks has stimulated significant improvements to monocular depth estimation as a supervised regression method~\cite{Liu01,Eigen01,Laina01}.
Recently, unsupervised training methods have been actively investigated. \cite{Godard01} used predicted disparity to synthesize a virtual image and minimized its photometric loss for training. \cite{Zhou01} trained the depth network jointly with an additional pose network, requiring only monocular sequences. Based on these approaches, they have widely been tackled~\cite{Godard02, Bian01, Xu2019RegionDN}. Many researchers have made further improvements along multiple lines, such as regularizing consistency with optical flow~\cite{Tosi2020DistilledSF, dfnet, CC, Zhao2020TowardsBG} or functional geometric constraints between feature maps~\cite{Spencer2020DeFeatNetGM, Shu2020FeaturemetricLF}.

Several recent works have proposed self-supervised depth prediction with semantics. They have enforced cross-task consistency and smoothness~\cite{Chen01, Ramirez01, zhu2020edge} and removed dynamic objects~\cite{Klingner2020SelfSupervisedMD} or explicitly modeled object motions~\cite{lee2021learning, wild, casser2018depth}. \cite{Guizilini2020SemanticallyGuidedRL} targeted semantics-aware representations for depth predictions, enabling it via knowledge transfer from a fixed teacher segmentation network with pixel-adaptive convolution~\cite{Su2019PixelAdaptiveCN}. In contrast, we design a multi-task network with cross-task multi-embedding attention and semantics-guided triplet loss to successfully produce semantics-aware representation.

\subsection{Neural Attention Network} 
\cite{attention,bert} designed a self-attention scheme that captures long-range dependencies to resolve the locality of recurrent operations. They proposed multi-head attention for utilizing information from different representation subspaces.
Recently, cross-attention schemes have been utilized to extract features across heterogeneous representations, such as image, speech, and text~\cite{hou2019cross, Wei_2020_CVPR, ZhaoNLJCM20b}. Additionally, for self-supervised depth estimation, \cite{Johnston2020SelfSupervisedMT, Zhou2019UnsupervisedHD} applied self-attention to capture the global context for estimating depth and combining multi-scale features from dual networks. In this study, inspired by the use of multi-head attention and cross attention, we propose a novel method of judiciously utilizing cross-task features across depth and segmentation.

\subsection{Multi-task Architecture}
The combination of features from multiple tasks has been widely used in recent multi-task architectures. \cite{Xu2018PADNetMG, mtinet, ECCV18} applied a convolutional layer to extract local information from the reference task feature for multimodal distillation. \cite{Zhou2020PatternStructureDF, zhang2019patternaffinitive, Jiao2019GeometryAwareDF, choi2020safenet} adopted affinity-guided message passing to propagate the relationship of spatially different features within the reference task to the target one. Instead, we propose a cross-task attention to fully utilize cross-modal interactions between geometry and semantics.
\subsection{Deep Metric Learning} 
Deep metric learning~\cite{Wang_2017_ICCV, song2017deep, MLarticle} has been widely applied in various fields, such as face recognition~\cite{Schroff_2015, Hu_2014_CVPR} and image ranking~\cite{wang2014learning, ng2020solar, netvlad}. Inspired by recent successes, we propose a semantics-guided triplet loss to refine feature representations for improving depth predictions by exploiting implicit geometry from semantic supervision.

\begin{figure*}[t]
    \centering
    \includegraphics[width=0.8\textwidth]{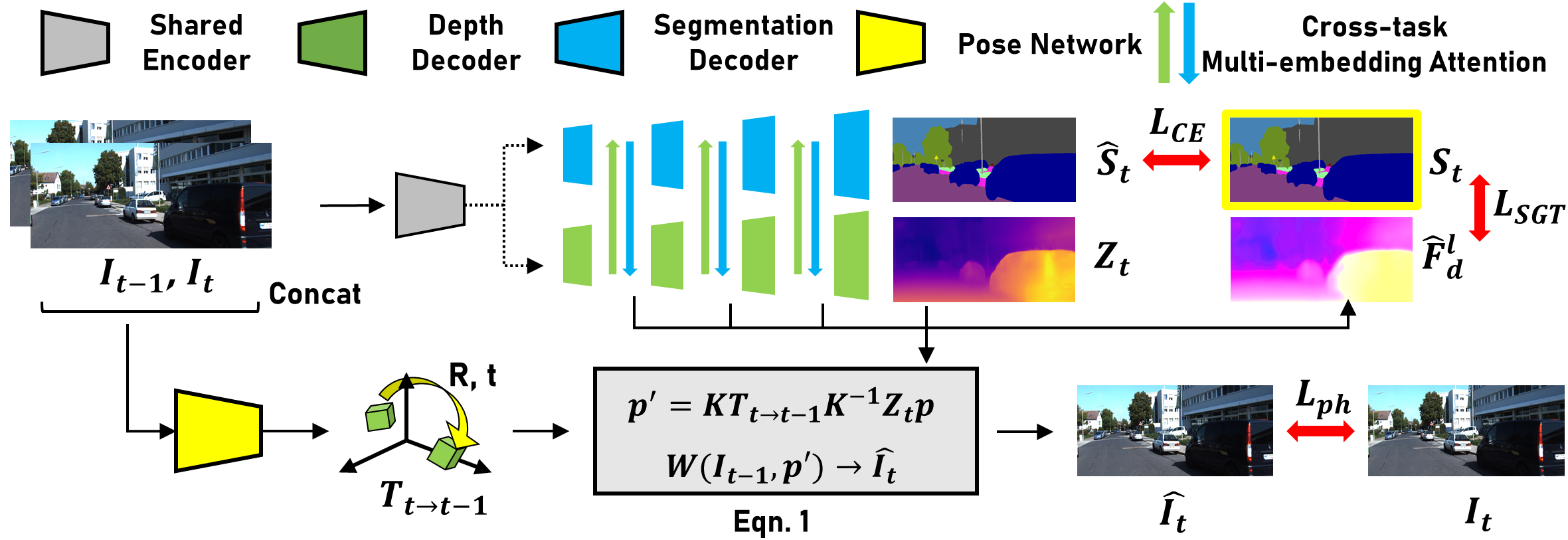}
    \vspace{-3mm}
    \caption{Overview of our proposed architecture. We adopt a shared encoder and two separate decoders, each for depth and segmentation, and we train both tasks jointly. The semantics-guided triplet loss ($L_{SGT}$) is calculated with the segmentation pseudo-label ($S_t$), and it optimizes L2 normalized depth feature maps ($\hat{F}^l_d$) with semantic guidance. The cross-task multi-embedding attention module is located between the layers of each decoder, and it enables cross-modal interactions between two task-specific decoders.}
    \vspace{-1.5mm}
    \label{fig:overview}
\end{figure*}

\section{Methods}
Here, we review our baseline approach, Monodepth2~\cite{Godard02}, and present our current methodology in the following subsections.
\subsection{Depth Estimation and Semantic Segmentation}
\label{preliminary}
\subsubsection{Self-supervised Monocular Depth Estimation}

Given consecutive RGB images, $I_{t-1}$ and $I_{t}$, one can predict $Z_t$, the depth of every pixel on $I_t$, and compute a six degree-of-freedom relative pose, $T_{t \to t-1}$, using a pose network.
With known camera intrinsics, $K \in \mathbb{R}^{3\text{x}3}$, we can derive the projected pixel coordinates and use them from $I_{t-1}$ as:
\begin{align}
    p' &= KT_{t \to t-1} Z_t K^{-1} p &
    \hat{I_t} &= W_t(I_{t-1}, p'), \label{eq:warp}
\end{align}
where $p$ is the homogeneous coordinates of the pixel in $I_t$, and $p'$ is the transformed coordinates of $p$ by $T_{t \to t-1}$. $W(\cdot)$ is a sub-differentiable bilinear sampler~\cite{Jaderberg2015SpatialTN} that obtains nearby pixels at $p'$ in $I_{t-1}$ and assigns the linearly interpolated pixel at $p$ in $\hat{I_t}$. 
Ideally, $I_t$ and $\hat{I_t}$ should be aligned if both depth and pose networks are optimally trained.
These two networks are jointly optimized to minimize the discrepancy between $I_t$ and $\hat{I_t}$. We utilize the structural similarity index measure (SSIM)~\cite{ssim} combined with L1 loss as a photometric loss, $L_{ph}$~\cite{Godard01}:
\begin{equation}
\label{loss:photometric}
    L_{ph} = \alpha \frac{1-\text{SSIM}(I_t, \hat{I_t})}{2} + (1 - \alpha)|I_t - \hat{I_t}|
\end{equation}

We compute $L_{ph}$ for the two frame pairs, $[I_{t-1}$, $I_{t}]$ and $[I_{t}$, $I_{t+1}]$ to deal effectively with occlusions. We apply the minimum reprojection~\cite{Godard02}, which selects the pixel having a smaller loss between the two reference frames $[I_{t-1}, I_{t+1}]$, and we apply an auto-mask~\cite{Godard02}. The following edge-aware smoothness loss~\cite{Godard01}, $L_{sm}$, is also added.
\begin{equation}
\label{loss:smoothness}
    L_{sm} = |\partial_x d_t|e^{-|\partial_xI_t|} + |\partial_y d_t|e^{-|\partial_yI_t|}
\end{equation}
The loss function of our baseline is obtained as follows:

\begin{equation}
\label{loss:baseline}
    L_{baseline} = L_{ph} + \beta \cdot L_{sm}
\end{equation}
where $\beta$ controls the relative strength of the smoothness factor.

\subsubsection{Supervised Semantic Segmentation} 
A typical network model for semantic segmentation has an encoder-decoder architecture~\cite{ronneberger2015unet} for extracting features and upsampling them for dense predictions. This structure is similar to our baseline depth network~\cite{Godard02}, wherein basic features are extracted first prior to being fed into the decoder. Therefore, we adopt a shared-encoder architecture to reduce computations and benefit from both tasks.

In our proposed method, we train semantic segmentation with a pseudo-label generated by an off-the-shelf segmentation model~\cite{Zhu01}. 
We do not require per-image ground-truth of segmentation in the training dataset; thus, it is more practically applicable. We used the cross-entropy loss, $L_{CE}$, for training, and the training loss includes $\gamma L_{CE}$ with the baseline loss (Eqn.~\ref{loss:baseline}), where $\gamma$ is a control parameter.

\begin{figure}[t]
     \centering
    \includegraphics[width=0.9\columnwidth]{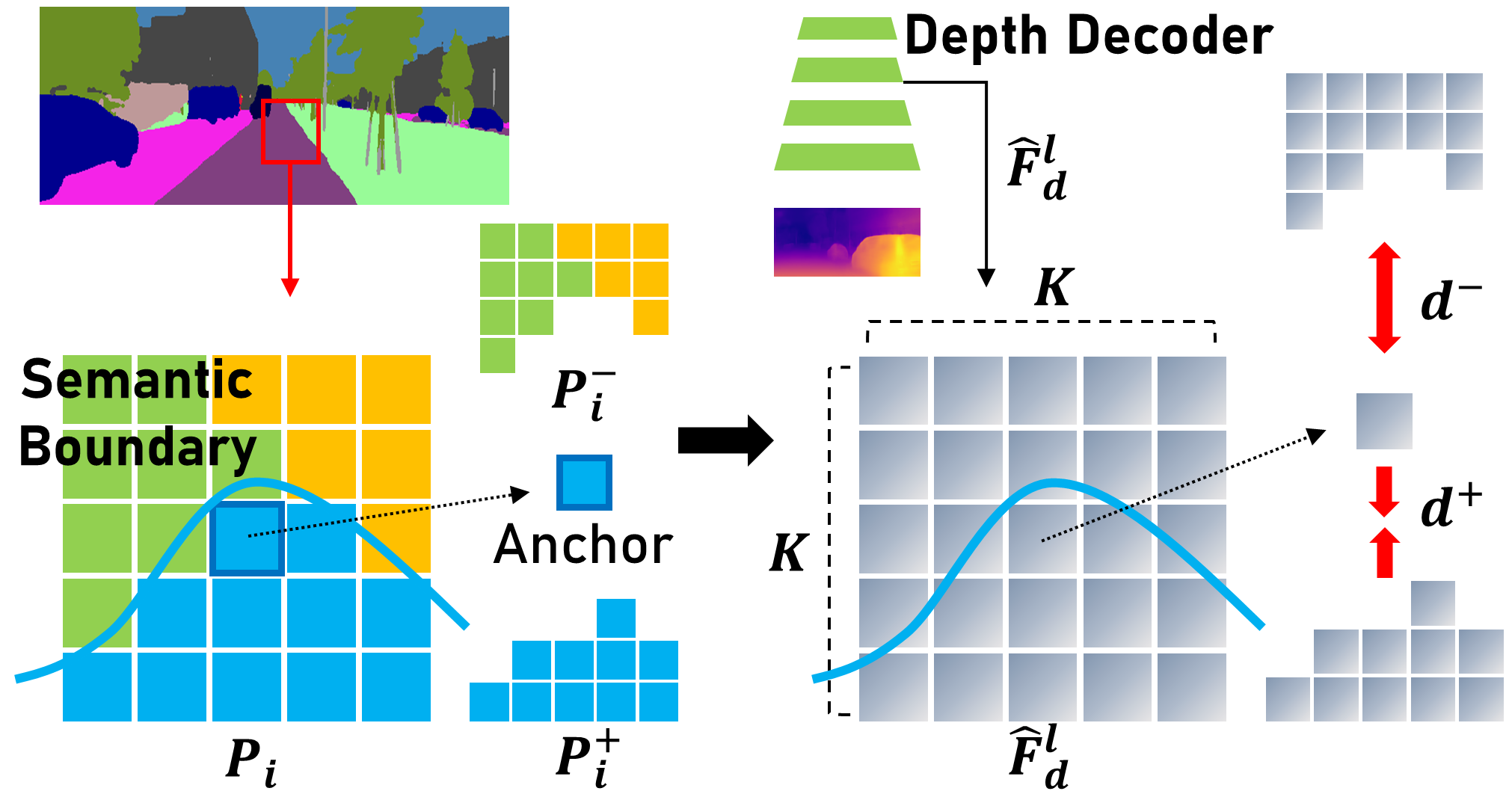}
    \vspace{-2mm}
    \caption{Overview of proposed semantics-guided triplet loss. We first divide $K \times K$ size local patch ($\mathcal{P}^i$) of the semantic label into triplets. Then, we optimize the distance between normalized depth features ($\hat{F}^l_d$ ) following the corresponding pixel locations in the semantic patch.}
 \vspace{-1.5mm}    
\label{fig:metric}
\end{figure}

\subsection{Semantics-guided Triplet Loss}
Based on the local geometric relation from scene semantics, adjacent pixels within each object instance have similar depth values, whereas those across semantic boundaries may have large depth differences. Thus, we apply this intuition through a representation learning problem inspired by the recent usage of deep metric learning~\cite{Wang_2017_ICCV, MLarticle}.
We first separate pixels of the local patch on the semantic label into triplets (i.e., anchor, positive, and negative), and we then divide features from the $l^{th}$ layer of the depth decoder ($F^l_d$) in accordance with the corresponding location of those triplets. We aim to optimize the distance among these triplets, following the intuition described above. However, we do not directly optimize the depth value itself. Our key idea is that the distance should be defined and optimized in the representation space. Hence, the depth decoder can produce more discriminative features on the boundary regions so that the output depth map becomes more aligned with the semantic boundaries.

\subsubsection{Patch-based Candidate Sampling}
\label{sec:3.3.1}
We first divide the semantic label into the $K$$\times$$K$ size of image patches with a stride of one. For each patch, we selected center of each patch as the anchor pixel and those that have the same class as that of the anchor as positive pixels. The negative pixels have different classes from those of the anchor pixels. Subsequently, we define ${\mathcal{P}_i}^+$ and ${\mathcal{P}_i}^-$, the sets of positive and negative pixels in the local patch $\mathcal{P}_i$, of which the spatial location of the anchor is $i$. We use ${\mathcal{P}_i}^+$ and ${\mathcal{P}_i}^-$ to determine whether $\mathcal{P}_i$ intersects the semantic borders. For example, $|{\mathcal{P}_i}^-| = 0$ means that $\mathcal{P}_i$ is located inside a specific object and does not cross the borders. On the other hand, if $|{\mathcal{P}_i}^+|$ and $|{\mathcal{P}_i}^-|$ are both larger than zero, it indicates that $\mathcal{P}_i$ intersects the boundaries across objects. 

Additionally, the semantic labels may not be accurate or consistent because they are predictions of pre-trained segmentation networks. To reduce misclassification caused by these imperfect labels, we set a threshold, $T$, and determine $\mathcal{P}_i$ intersects with the boundaries when $|{\mathcal{P}_i}^+|$ and $|{\mathcal{P}_i}^-|$ are both larger than $T$.

\subsubsection{Triplet Margin Loss}
We grouped the features in each patch of the depth feature map into three classes (i.e., anchor, positive, and negative) following the corresponding pixel locations in the semantic image patch. We define positive distance $d^+$ and negative distance $d^-$ as the mean of the Euclidean distance of the L2 normalized depth feature pairs.
\begin{align}
    & d^+(i) = \frac{1}{|{\mathcal{P}^+_i}|}\sum_{j \in \mathcal{P}^+_i}{\sqrt{(\hat{F^l_d}(i) - \hat{F^l_d}(j))^2}}  \\ 
    & d^-(i) = \frac{1}{|{\mathcal{P}^-_i}|}\sum_{j \in \mathcal{P}^-_i}{\sqrt{(\hat{F^l_d}(i) - \hat{F^l_d}(j))^2}},  
\end{align}
where $\hat{F^l_d} = F^l_d / \|{F^l_d}\|$.

We aim to reduce the distance between the anchor and positive features, and increase the distance between the anchor and negative features. However, naively maximizing $d^-$ as far as possible does not lead to our desired outcome because the semantic border does not always guarantee that the depth of two separate objects differs by a large amount. Instead, we adopt the triplet loss~\cite{triplet, wang2014learning} with a margin so that the distance is no longer optimized when the negative distance exceeds a positive distance more than a specific margin $m$, as a hyper-parameter.
\begin{align}
    \mathcal{L}_{\mathcal{P}_i} = max(0, d^+(i) + m - d^-(i)) 
\end{align}
The semantics-guided triplet loss $L_{SGT}$ is the average of $\mathcal{L}_{\mathcal{P}_i}$, only containing $\mathcal{P}_i$ satisfying the condition described in Sec.~\ref{sec:3.3.1}.
\begin{align}
    \mathcal{L}_{SGT} = \frac{\sum_{i}{\mathbbm{1}{[ |{\mathcal{P}_i}^+|,|{\mathcal{P}_i}^-|>T ]}\cdot{\mathcal{L}_{\mathcal{P}_i}}}}{ \sum_{i}{\mathbbm{1}{[ |{\mathcal{P}_i}^+|,|{\mathcal{P}_i}^-|>T ]}}}
\end{align}
We sum over the $L_{SGT}$ of depth features across multiple layers and include into the total loss the sum multiplied by control parameter $\delta$.

\subsection{Cross-task Multi-embedding Attention (CMA) Module}
We propose a CMA module to produce semantics-aware depth features through the representation subspaces and utilize them to refine depth predictions. As illustrated in Fig.~\ref{fig:overview}, the CMA modules are located in the middle of each decoder layer and utilize the information from the other decoder. A single CMA module has uni-directional data flow, e.g., a CMA module refines the target feature with the reference feature. We use two CMA modules simultaneously to enable bidirectional feature enhancement, where depth (segmentation) becomes the target (reference) in one CMA module while their roles change in the other. In the following paragraphs, we only describe a single case where the depth feature is the target for ease of explanation. 

In our model, each decoder comprises five blocks ($l=0,1,2,3,4$) and the spatial resolution of the feature map is doubled for each. The depth (segmentation) decoder generates a feature map, $F^l_d$ ($F^l_s$), which has a spatial resolution of $(H / 2^{4 - l}, W / 2^{4 - l})$, where $H$ and $W$ are the height and width of the input image, respectively. The CMA modules can be attached to any of the five candidates.

The CMA module performs a pixel-wise operation on the two feature maps, $F^l_d$ and $F^l_s$, through several operations. It first computes the semantic awareness of the depth features as a pixel-wise attention score through cross-task similarity (Sec.~\ref{CMA:sim}). We then extend this computation with multiple linear projections so that the similarity can be computed from different representation subspaces (Sec.~\ref{CMA:atten}). This enables selective extraction of depth features from multiple embeddings upon the corresponding semantic awareness, maximizing the utilization of cross-modality. Subsequently, the fusion function combines the input feature map, $F^l_d$, with the refined one, ${F^l_d}'$ (Sec.~\ref{CMA:fusion}). We explain the details in the following sections.

\subsubsection{Cross-task Similarity} \label{CMA:sim}
We define {\it cross-task similarity} as $F^l_d(i)^T F^l_s(i)$, where $i$ is the spatial index of each feature map, and $F^l(i)$ is a {\it $C$-dimensional} feature vector. This indicates quantitative amounts of semantic representation that each depth feature implicitly refers to. However, direct computation with raw feature vectors is infeasible, owing to the different nature of the tasks. We apply a linear projection, $\phi$, that transforms the input feature from the original dimension, $C$, to $C'$. This indirectly computes the cross-task similarity through the representation subspace. The refined feature is computed as follows:
\begin{align}
    & {F^l_d(i)}' = \rho(A(i))\times\phi_v(F^l_d(i)),\label{eq:cross-task similarity_1}\\ 
    & \text{where} \quad A(i) = \frac{\phi_k(F^l_d(i))^T\phi_q(F^l_s(i))}{ \sqrt{C'}} \label{eq:cross-task similarity_2}
\end{align}
Here, $\rho$ is a normalization factor scaling the input. We apply three separate linear embeddings, and each acts as query ($\phi_q$), key ($\phi_k$), and value ($\phi_v$) functions. The target feature map, $F^l_d$, becomes the input for the key and value embeddings, and the reference feature map, $F^l_s$, becomes the input for the query embedding.

For depth prediction, this imposes large attention scores ($A(i)$) on the specific depth features which are consistent with semantics, so that it can implicitly utilize semantic region information. As mentioned above, this module is bidirectional, and the semantic feature $F^l_s$ acts as the target simultaneously. At this time, the depth feature is used to learn the features for semantic prediction so that backpropagation from the segmentation loss ($L_{CE}$) optimizes depth layers while offering more semantics-aware representations.

Compared with the affinity matrix for cross-task feature fusion~\cite{Zhou2020PatternStructureDF, zhang2019patternaffinitive, Jiao2019GeometryAwareDF, choi2020safenet}, which is computed solely based on features from a single task, the CMA module computes the attention score based on features from both tasks. Hence, it can effectively handle cross-modal interactions for multi-task predictions.

\begin{figure}[t]
     \centering
    \includegraphics[width=0.9\columnwidth]{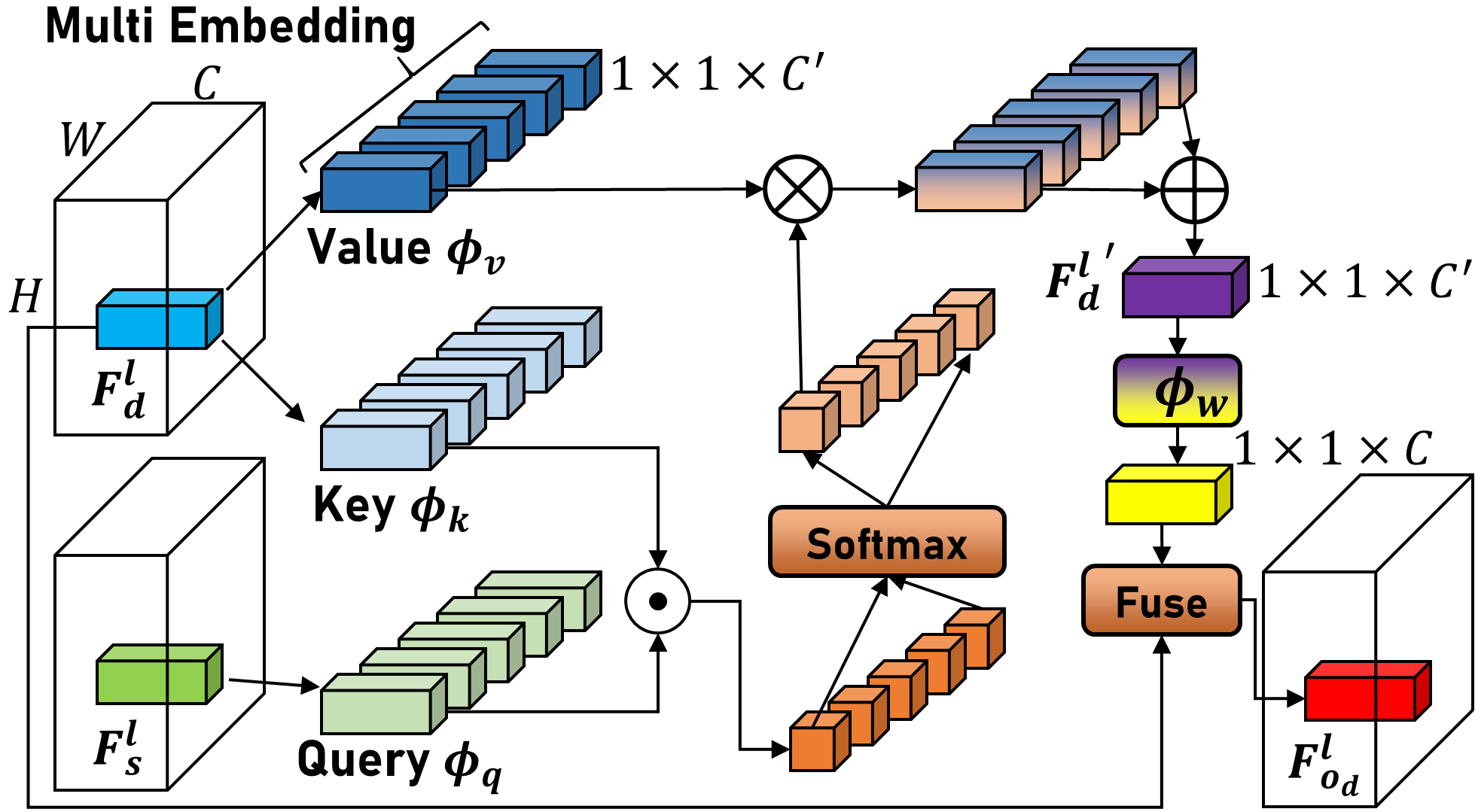}
    \vspace{-2mm}
    \caption{Architecture of the cross-task multi-embedding attention module. The feature map from the $l^{th}$ layer of the depth decoder, $F^l_d$, acts as a target in this case, and the segmentation feature, $F^l_s$, acts as a reference. The CMA module refines the target feature with the reference feature, and produces the output feature $F^l_{o_d}$.} 
\label{fig:att}
    
\end{figure}

\subsubsection{Multi-embedding Attention} \label{CMA:atten}
Inspired by multi-head attention~\cite{attention, bert}, we adopt multiple linear projections to compute the similarity between feature vectors through different representation subspaces. This refines depth features with implicit semantic representations more effectively, as verified in Sec.~\ref{section:exp}. We use $H$ distinct projection functions, $\phi^h (h = {1,...,H})$; hence, the queries, keys, and values are mapped to $H$ independent subspaces. The cross-task similarity in Eqs.~\ref{eq:cross-task similarity_1}-\ref{eq:cross-task similarity_2} can be directly extended to a multi-embedding scheme as follows:
\begin{align}
    & {{F^l_d}^h(i)}' = \rho({A^h(i)})\times\phi^h_v(F^l_d(i)),\\ \label{multi-embedding}
    & \text{where} \quad \rho(A^h(i)) = \frac{e^{A^{h}(i)}}{\sum_{h'\in H}{e^{A^{h'}(i)}}}, \\
    & \text{and} \quad A^h(i) = \frac{\phi^h_k(F^l_d(i))^T\cdot\phi^h_q(F^l_s(i))}{\sqrt{C'}}
\end{align}

The refined feature, ${{F^l_d}(i)}'$, is the summation of the feature maps refined from each embedding function:
\begin{align}
    & {{F^l_d}(i)}' = \sum_{h}{{{F^l_d}^h(i)}'} 
\end{align}
In the above equations, $h$ represents the index of multiple linear embeddings. We adopted $softmax$ as a normalization function, $\rho$, to compute the importance of each embedding. Thus, we can selectively exploit the outputs from multiple attentions. This process is illustrated in Fig.~\ref{fig:att}.

In contrast to the original multi-head attention where the results from each embedding head are concatenated and equally handled, we compute the attention score among multiple heads and measure the significance of results from each embedding on the corresponding attention scores. 


\subsubsection{Fusion Layer} \label{CMA:fusion}
Finally, the refined feature map, ${F^l_d}'$, is projected to the original dimension $C$ ($\phi_w$ in Fig.~\ref{fig:att}) and fused with the initial feature map, $F^l_d$, to produce final output, $F^l_{o_d}$. 
We apply two convolution layers to concatenated feature maps, $[\phi_w({F^l_d}'), F^l_d]$, to produce $F^l_{o_d}$. $F^l_{o_d}$ becomes the input of ${l+1}^{th}$ layer of the depth decoder.

\section{Experiments}
\label{section:exp}

\subsection{KITTI Dataset}
The KITTI dataset~\cite{Geiger01} has been widely adopted for depth prediction benchmarks. We used the Eigen split~\cite{Eigen01} for this purpose, and preprocessing was performed to remove static frames, as in \cite{Godard02,Zhou01}; thus, 39,910 and 4,424 images were used for training and validation, respectively, and 697 images were used for evaluation. 

For training semantic segmentation, we generated pseudo-labels using an off-the-shelf network~\cite{Zhu01}. To evaluate the segmentation performance, we used 200 images and labels provided in the training set of the KITTI semantic segmentation benchmark corresponding to KITTI 2015~\cite{Menze2015CVPR}. 

\begin{table*}[t]
\footnotesize
\centering
\begin{subtable}[c]{2\columnwidth}
\centering
 \begin{threeparttable}
\begin{tabular}{c | c  | c  || c  c  c  c | c  c  c }
\toprule
 & & & \multicolumn{4}{c|}{\underline{Lower is better}}  & \multicolumn{3}{c}{\underline{Higher is better}} \\
 Method & Backbone & Sem & AbsRel & SqRel & RMS & RMSlog & $< 1.25$ & $< 1.25^2$ & $< 1.25^3$  \\
\midrule
SceneNet~\cite{Chen01} & DRN~\cite{yu2017dilated} & \checkmark& 0.118 & 0.905 & 5.096 & 0.211 & 0.839	& 0.945 & 0.977  \\
\midrule
Monodepth2~\cite{Godard02}~\tnote{*} & ResNet18    &  & 0.114 & 0.864 & 4.817 & 0.192 & 0.875	& 0.959 & 0.981  \\
Zou {\it et al.}~\cite{zou2020learning}&ResNet18    & & 0.115 & 0.871 & 4.778 & 0.191 & 0.874 & 0.963 & 0.984  \\
Guizilini {\it et al.}~\cite{Guizilini2020SemanticallyGuidedRL}&ResNet18  &\checkmark& 0.117 & 0.854 & 4.714 & 0.191 & 0.873 & 0.963 & 0.981 \\
SGDepth~\cite{Klingner2020SelfSupervisedMD}&ResNet18 &\checkmark & 0.113 & 0.835 & 4.693 & 0.191 & 0.879 & 0.961 & 0.981 \\
Lee {\it et al.}~\cite{lee2021learning}&ResNet18 &\checkmark & 0.112& 0.777 &4.772& 0.191 &0.872& 0.959& 0.982 \\
Poggi {\it et al.}~\cite{poggi2020uncertainty} &ResNet18 &  & 0.111 & 0.863 & 4.756 & 0.188 & 0.881 & 0.961 & 0.982   \\
Patil {\it et al.}~\cite{Patil2020DontFT}&ResNet18  &  & 0.111 & 0.821  & 4.650 & 0.187 &  0.883 & 0.961 & 0.982 \\
SAFENet~\cite{choi2020safenet}&ResNet18 & \checkmark &  0.112 & 0.788 & 4.582 & 0.187 & 0.878 & 0.963 & 0.983 \\
HRDepth~\cite{lyu2020hrdepth}&ResNet18 & & 0.109 & 0.792 & 4.632 & 0.185 & 0.884 & 0.962 & 0.983  \\

{\bf Ours}&ResNet18 &\checkmark & {\bf 0.105} &	{\bf 0.722} &  {\bf 4.547}	& {\bf 0.182}	& {\bf 0.886} & {\bf0.964} &	{\bf 0.984}
  \\ 

\midrule
SGDepth~\cite{Klingner2020SelfSupervisedMD} & ResNet50 &\checkmark & 0.112 & 0.833 & 4.688 & 0.190 & 0.884 & 0.961 & 0.981 \\
Monodepth2~\cite{Godard02} & ResNet50  &  & 0.110 & 0.831 & 4.642 & 0.187 & 0.883 & 0.962 & 0.982  \\
Guizilini {\it et al.}~\cite{Guizilini2020SemanticallyGuidedRL}& ResNet50  &\checkmark&0.113 & 0.831 & 4.663 &  0.189 & 0.878 & {\bf 0.971} & 0.983  \\
FeatDepth~\cite{Shu2020FeaturemetricLF}& ResNet50 & & 0.104 & 0.729 & 4.481 & 0.179 & {\bf 0.893} & 0.965 & 0.984   \\
Li {\it et al}~\cite{li2021learning} & ResNet50  & \checkmark & 0.103 & 0.709 & 4.471  & 0.180 &  0.892  & 0.966  & 0.984 \\
{\bf Ours} & ResNet50  &\checkmark &   {\bf 0.102} &	{\bf 0.675} &	{\bf 4.393}  &	{\bf 0.178} &	{\bf 0.893}	&  0.966 &	{\bf 0.984}	
  \\
\midrule
Johnston {\it et al}~\cite{Johnston2020SelfSupervisedMT} & ResNet101  &  & 0.106 & 0.861 & 4.699 & 0.185 & 0.889 & 0.962 & 0.982  \\
\midrule
PackNet-SfM~\cite{Guizilini20203DPF} & PackNet  & &  0.111 & 0.785 & 4.601 & 0.189 & 0.878 & 0.960 & 0.982 \\
Guizilini {\it et al.}~\cite{Guizilini2020SemanticallyGuidedRL}& PackNet  &\checkmark& 0.102& 0.698& 4.381 &0.178& 0.896 &0.964 &0.984 \\

\bottomrule
\end{tabular}
  \begin{tablenotes}
  \item[*] We re-implemented Monodepth2 and the result has slightly improved from the original paper.
  \end{tablenotes}
 \end{threeparttable}
\vspace{-2mm}
\caption{}
\label{table:sota}

\end{subtable}
\begin{subtable}[c]{2\columnwidth}
\centering

\begin{tabular}{c | c | c | c || c  c  c  c | c  c  c }
\toprule
 $L_{CE}$ & $L_{SGT}$ & CMA module &  HR & AbsRel & SqRel & RMS & RMSlog & $< 1.25$ & $< 1.25^2$ & $< 1.25^3$  \\
\midrule
  & & & & 0.114	& 0.864 & 4.817 & 0.192 & 0.875	& 0.959 & 0.981 \\ 
  \checkmark &  & & & 0.112 & 0.823 & 4.705 & 0.189 & 0.879 & 0.961 & 0.982   \\ 
  \checkmark  & \checkmark   & & &  0.108 & 0.755	& 4.618	& 0.186	&0.882	& 0.962	& 0.983	\\ 
  \checkmark  &  &  \checkmark         & & 0.107	& 0.741	& 4.586	& 0.184	& 0.884	& 0.962	& 0.983 \\
  \checkmark & \checkmark & \checkmark & &  {\bf 0.105} & {\bf 0.722} &  {\bf 4.547}	&  {\bf 0.182}	&  {\bf 0.886} & {\bf 0.964} &	 {\bf 0.984}
  \\ 
  \midrule
  \checkmark & \checkmark & \checkmark  & \checkmark & 0.102 &	0.687 &	4.366 &	0.178 &	0.895 &	0.967 &	0.984 \\
\bottomrule
\end{tabular}
\vspace{-2mm}
\caption{}
\label{table:ablation}

\end{subtable}
\vspace{-3mm}
\caption{(a) Comparison of self-supervised monocular depth estimation with recent works on KITTI Eigen split. All methods are trained with monocular images with size of 192$\times$640, except SceneNet. Sem denotes training with semantic information. (b) Ablation of the proposed methods in depth predictions. HR denotes training with high-resolution images (320$\times$1024). }
 \vspace{-1.5mm}
\end{table*}

\subsubsection{Evaluation}
To evaluate the capability of depth prediction, we conducted per-image median-scaling with ground-truth following the evaluation protocol in \cite{Godard02}. The maximum depth is 80 m, as in recent studies~\cite{Godard02,Patil2020DontFT,Klingner2020SelfSupervisedMD}. We evaluated semantic segmentation in the mean intersection over union (mIoU).

\begin{table}[ht]
\footnotesize
\centering
\begin{tabular}{c | c || c   }
\toprule
Method & Train & MIoU \\
\bottomrule
SceneNet~\cite{Chen01} & CS & 37.7 \\
SGDepth~\cite{Klingner2020SelfSupervisedMD} & CS & 51.6  \\
Ours & K & 56.6 \\
{\bf Ours (HR)} & K & {\bf 60.6} \\
\midrule
Ours w/o CMA & K & 56.1 \\
Ours w/o CMA (HR) & K & 59.1 \\
\bottomrule
\end{tabular}
\vspace{-2mm}
\caption{Semantic segmentation results on KITTI 2015 training set. CS denotes Cityscapes, and K represents KITTI. HR refers to training using high-resolution image.}
\vspace{-1.5mm}
\label{table:semantic}

\end{table}

\subsection{Implementation Details}
\subsubsection{Network Architecture}
The depth and segmentation network has a standard encoder-decoder architecture~\cite{ronneberger2015unet} with skip connections, as in Monodepth2~\cite{Godard02}. The shared encoder and the pose network encoder are ResNet-18~\cite{resnet}, pre-trained on ImageNet~\cite{imagenet_cvpr09}. For the CMA module, we adopt four ($H$ = 4) embeddings of the multi-embedding scheme. The dimension ratio between the original feature and the embedded feature is two, such that $C'$ = $2*C$. Thus, the projected vectors have twice the dimensions of the the corresponding input features. The normalization factor, $\rho$, is the identity function when $H=1$ (without multi-embedding) and $softmax$ when $H>1$ (w/ multi-embedding). We apply CMA module to three of decoder layers, $l=0,1,2$. 

\subsubsection{Training Details} For training, we resized the original image into a resolution of $192\times640$ and used a batch size of 12. The Adam optimizer was used with an initial learning rate of 1.5e-4, and we trained for 20 epochs while the learning rate was decayed by 0.1 twice, after 10 and 15 epochs of training. We used SSIM~\cite{ssim} with $L_1$ loss for $L_{ph}$, with $\alpha$ = 0.85 following the previous work~\cite{Godard02}.
We set the loss parameters as follows: $\beta=0.001$, $\gamma = 0.3$, and  $\delta = 0.1$. The local patch size, $K$, is set to five, and the margin, $m$, is set to 0.3 for $L_{SGT}$. This loss is applied to features from three layers, $l=1,2$ and $3$. The threshold $T$ is set to $K-1$.

\subsection{Quantitative Results and Ablation Study}
Table~\ref{table:sota} compares our proposed method with recent works. Ours achieves state-of-the-art results on the KITTI Eigen test split and outperforms previous works in every metric. Our network adopts ResNet-18 as a backbone, but we also use ResNet-50 and compare it with others adopting ResNet-50. Ours (ResNet-50) also achieved the best results. Note that the PackNet versions of ~\cite{Guizilini2020SemanticallyGuidedRL} adopted a significantly large backbone ($>10\times$ larger than ResNet-18). Therefore, we compare the ResNet-18 and ResNet-50 versions of \cite{Guizilini2020SemanticallyGuidedRL} and show that our method outperforms it by a large margin.
Additionally, in our multi-task network, semantic information is required only for training. In contrast, \cite{Guizilini2020SemanticallyGuidedRL} and \cite{li2021learning} require the semantics for both training and testing. \cite{Guizilini2020SemanticallyGuidedRL} requires a teacher segmentation network for feature distillation during inference and \cite{li2021learning} requires semantic label or pre-computed segmentation results as the network input. 
Finally, our network is highly compatible with more advanced networks~\cite{lyu2020hrdepth, Guizilini20203DPF} which have architectural differences from our baseline, Monodepth2. This indicates the potential for further improvement.

In Table~\ref{table:ablation}, we also evaluate the effectiveness of each proposed method. The addition of semantic segmentation to depth ($L_{CE}$) via shared encoder shows an improvement. Applying the semantics-guided triplet loss and the CMA module further improves the baseline. This verifies that more semantics-aware representation improvements of depth predictions are produced. Finally, the combination of both methods significantly improves the performance. Both techniques are designed to refine the depth representation via semantic knowledge, and they offer highly synergistic improvements.

In Table~\ref{table:semantic}, we also evaluate the semantic segmentation performance on KITTI 2015~\cite{Menze2015CVPR}. Though the proposed method outperforms others, it is not fair to compare with the works that trained semantic segmentation with Cityscapes~\cite{Cordts2016Cityscapes} ground-truth (they trained depth on KITTI.). Hence, we focus more on how segmentation benefits from depth estimation via CMA rather than the final performance. As shown in the last two rows, the proposed CMA module also improves the segmentation performance. Thanks to its bi-directional flow, the CMA module also refines semantic features as a target with reference to depth representations. Additionally, it is more effective when the resolution is high. 

\begin{figure*}[t!]
     \centering
     \footnotesize
     \begin{subfigure}[b]{0.25\textwidth}
         \centering
         \stackunder[1pt]{\includegraphics[width=\textwidth]{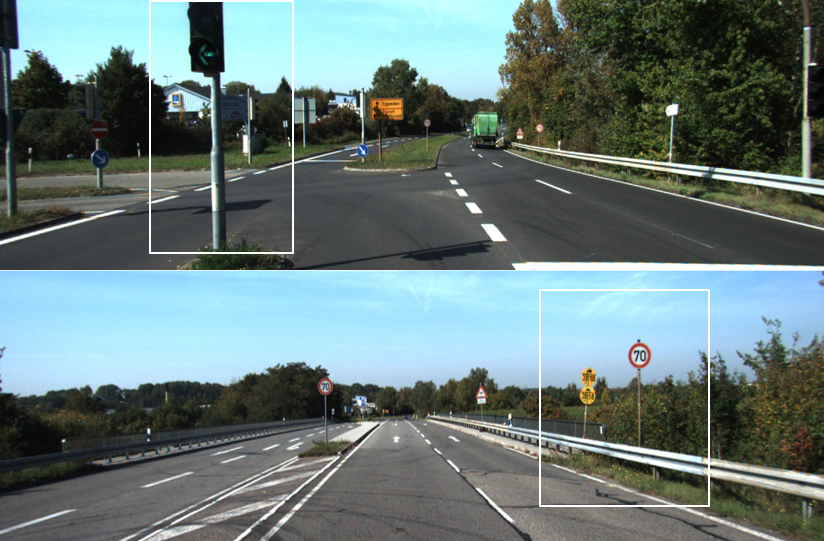}}{Image}%
     \end{subfigure}
\hspace*{-0.9em}
     \begin{subfigure}[b]{0.25\textwidth}
         \centering
         \stackunder[1pt]{\includegraphics[width=\textwidth]{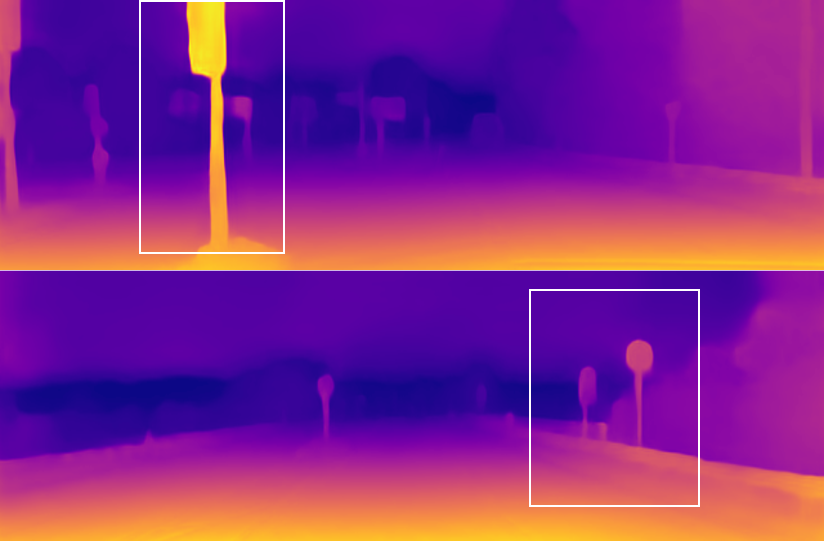}}{Ours}%
     \end{subfigure}
\hspace*{-0.9em}  
 \begin{subfigure}[b]{0.25\textwidth}
         \centering
         \stackunder[1pt]{\includegraphics[width=\textwidth]{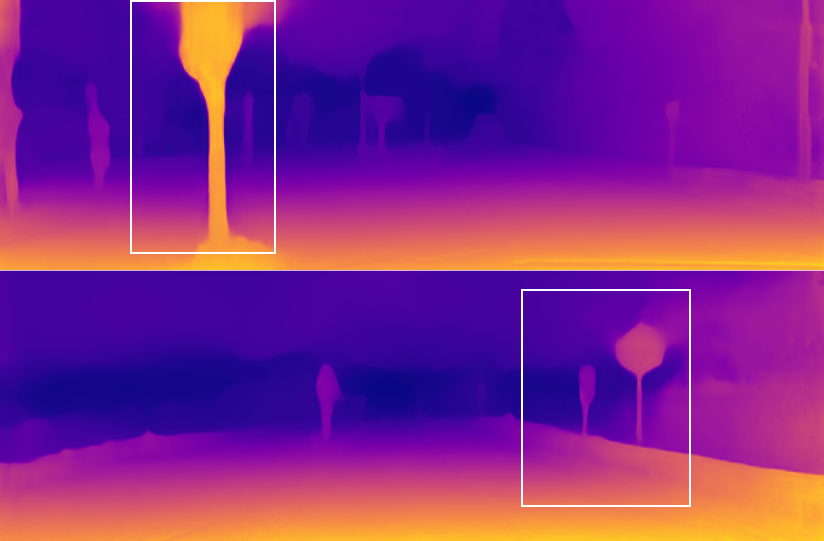}}{SGDepth~\cite{Klingner2020SelfSupervisedMD}}%
     \end{subfigure}
\hspace*{-0.9em}
 \begin{subfigure}[b]{0.25\textwidth}
         \centering
         \stackunder[1pt]{\includegraphics[width=\textwidth]{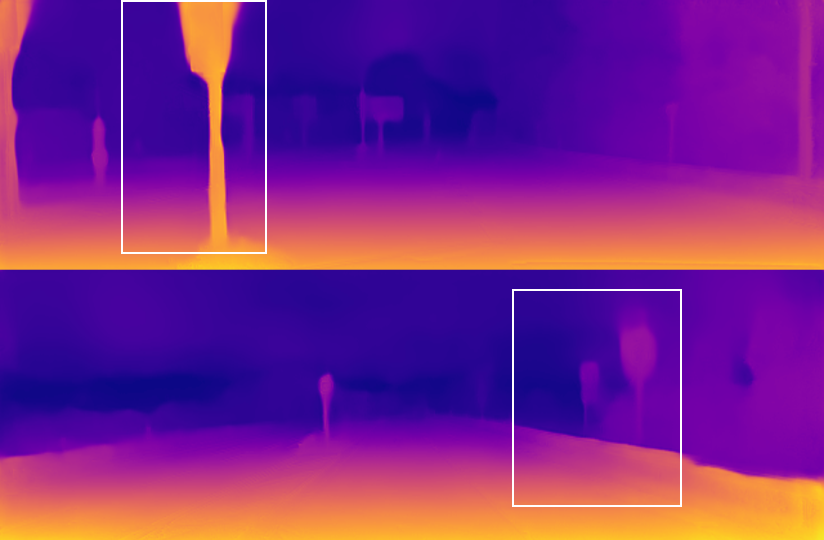}}{PackNet-Sfm~\cite{Guizilini20203DPF}}%
     \end{subfigure}
\vspace{-3mm}
        \caption{Qualitative comparison of the depth predictions with recent works.}
         \vspace{-1.5mm}
        \label{fig:visual}

\end{figure*}

\subsection{Qualitative Evaluation}
We qualitatively compare our method with recent methods, SGDepth~\cite{Klingner2020SelfSupervisedMD} and PackNet-Sfm~\cite{Guizilini20203DPF}, as shown in Figs.~\ref{fig:intro} and \ref{fig:visual}. In Fig.~\ref{fig:intro}, we compare the depth predictions and error distributions\footnote{Owing to the sparsity of depth ground-truth, we computed the mismatch with top-performing supervised depth network~\cite{lee2019big}.}, fixing AbsRel between 0 and 1. Similar to ours, \cite{Klingner2020SelfSupervisedMD} also adopted multi-task training with semantic segmentation via a shared encoder. However, only enhancing the encoder in a multi-task setting cannot fully exploit the semantic information. As shown in both figures, our method captures fine-grained detail, leading to more accurate depth predictions compared with others, especially at object borders. This verifies the effectiveness of the proposed fine-grained semantics-aware enhancement of representation.

\begin{table}[t!]
\footnotesize
\begin{subtable}[c]{.5\linewidth}
\centering
\begin{tabular}{c|| c | c | c  c || c | c | c   }
\toprule
 $l$ & AbsR & SqR & $<$$1.25$ & $K$ & AbsR & SqR & $<$$1.25$ \\
\bottomrule
 0-3&  0.108 & 0.757 & 0.882 & 3 & 0.109 & 0.757 & 0.878 \\
 1-3&  {\bf 0.108} & 0.755 & {\bf 0.882} & 5 & {\bf 0.108} & {\bf 0.755} & {\bf 0.882}\\
 1-4&  0.108 & {\bf 0.742} & 0.879 & 7 & 0.110 & 0.776 & 0.880\\
\bottomrule
\multicolumn{4}{c}{(a)} & \multicolumn{4}{c}{(b)} \\
\end{tabular}
\end{subtable}

\vspace{-3mm}
\caption{Ablations of semantics-guided triplet loss with (a) different layers of depth decoder the loss is applied, (b) the size of local patch $K$.}
 \vspace{-1.5mm}
\label{table:sgt}
\end{table}

\subsection{Further Analysis}
Table~\ref{table:sgt}a shows the results of varying layers ($l=0,1,2,3,4$) to which semantics-guided triplet loss is applied. We selected  layers $l=1,2$ and $3$ because it showed the best results. Applying $L_{SGT}$ to $l=4$ degrades the performance as it has a significantly low channel dimension (16); hence, the distance cannot be properly computed. In Table~\ref{table:sgt}b, we compare the effect of the patch size, $K$. Because separating a local patch into triplets relies on semantic labels from the off-the-shelf network, there must be noisy labels. When $K$ is small (i.e., $K=3$), the number of samples $|{\mathcal{P}^{+,-}}|$ decreases, and each noisy label contributes more to the mean distance, $d^{+,-}$. When $K$ is large (i.e., $K=7$), each local patch contains more non-boundary pixels and the negative distance can easily exceed the margin. In other words, the loss is computed from more {\it easy samples}, and the improvements are limited. In our experiments, $K=5$ was the balanced point, which was the best option. We further compare the results of different margins, $m$, in the supplement.

Table~\ref{table:cma} lists the results of the CMA module with varying parameters. As shown in Table~\ref{table:cma}a, our bidirectional CMA provides better results than the unidirectional CMA. This confirms that, to benefit from cross-modal representation, it is more beneficial to simultaneously improve both depth and semantics features than to improve just one.

\begin{figure}[t]
    \centering
    \includegraphics[width=0.95\columnwidth]{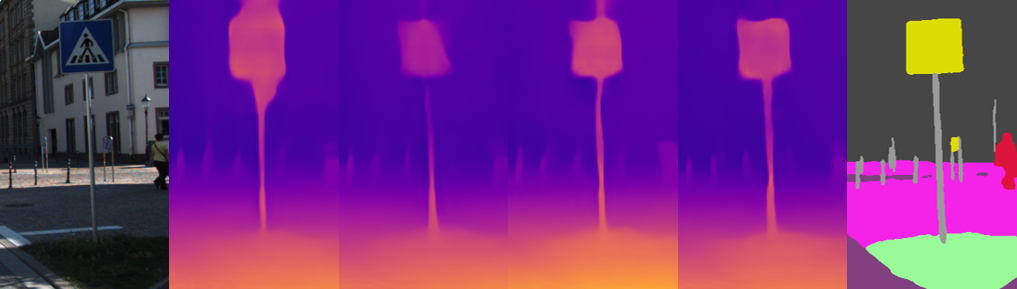}
    \vspace{-2mm}
    \caption{Comparison among baseline ($+L_{CE}$), and CMA module applied ($H=1,2,4$) from left to right.}
     \vspace{-1.5mm}
    \label{fig:ablation}
\end{figure}

Table~\ref{table:cma}b shows the effectiveness of the proposed multi-embedding scheme. It can fully utilize cross-modality as the number of embeddings grows. As shown in  Fig.~\ref{fig:ablation}, the more embeddings used, the more precise the object boundary of the depth network, and the depth prediction becomes more aligned to semantics. This demonstrates that the depth network can have more semantics-aware representations, owing to our proposed multi-embedding scheme.

\begin{table}[!t]
\footnotesize
\begin{subtable}[c]{.5\linewidth}
\centering
\begin{tabular}{c|| c | c | c  c || c | c | c   }
\toprule
 T & AbsR & SqR & $<$$1.25$ & $H$ & AbsR & SqR & $<$$1.25$ \\
\bottomrule
 D&  0.108 & 0.759 & 0.882 & 1 & 0.108 & 0.762 & 0.881 \\
 S&  0.111 & 0.797 & 0.877 & 2 & 0.107 & 0.749 & 0.882\\
 D,S&  {\bf 0.107} & {\bf 0.741} & {\bf 0.884} & 4 & {\bf 0.107} & {\bf 0.741} & {\bf 0.884}\\
\bottomrule
\multicolumn{4}{c}{(a)} & \multicolumn{4}{c}{(b)} \\
\end{tabular}
\end{subtable}
\vspace{-3mm}
\caption{Ablations of CMA module with (a) different target task features and (b) the number of embedding functions. T denotes the target task, D denotes the depth, and S denotes segmentation.}
 \vspace{-1.5mm}
\label{table:cma}
\end{table}

\section{Conclusion}
This paper proposed novel methods for accurate monocular depth prediction (i.e., semantics-guided triplet loss and cross-task multi-embedding attention) to make the best use of semantics-geometry cross-modality. Semantics-guided triplet loss offered a new and effective supervisory signal for optimizing depth representations. The CMA module allowed us to utilize rich and spatially fine-grained representations for multi-task training of depth prediction and semantic segmentation. The enhanced representation from these two methods exhibited a highly synergistic performance boost. Our extensive evaluation on the KITTI dataset demonstrates that the proposed methods outperformed extant state-of-the-art methods, including those that use semantic segmentation.
\\ \\
{\bf Acknowledgement.} We would like to especially thank Soohyun Bae at Bobidi for his invaluable comments. This work was supported by the SNU-SK Hynix Solution Research Center (S3RC).
{\small
\bibliographystyle{ieee_fullname}
\bibliography{main}
}

\end{document}